\documentclass{article}
\usepackage[utf8]{inputenc}

\usepackage[vlined,ruled,linesnumbered]{algorithm2e}
\usepackage{array}
\usepackage{amsmath}
\usepackage{color}
\usepackage{multirow}
\usepackage{caption}
\usepackage{subcaption}
\usepackage{graphicx}
\usepackage{fancyhdr}

\newcommand{\omitit}[1]{}

\fancypagestyle{fancyfirst}
{
\fancyhead{}
\fancyfoot[C]{
\footnotesize DISTRIBUTION STATEMENT A. Approved for public release. Distribution unlimited.\\
}
}

\title{Synthetically Generating Human-like Data for Sequential Decision Making Tasks via Reward-Shaped Imitation Learning}
\author{Bryan Brandt$^{a,b}$ and Prithviraj Dasgupta$^a$\\[0.05in]
$^a$Distributed Intelligent Systems Section\\
Information Technology Division\\
Naval Research Laboratory, Washington, D. C., USA\\[0.05in]
$^b$Computer Science Department\\
University of Wisconsin-Whitewater, WI, USA\\[0.05in]
E-mail: bcbrandt21@uww.edu, raj.dasgupta@nrl.navy.mil
}

\date{}

\begin{document}

\maketitle
\noindent
{\bf Abstract.} We consider the problem of synthetically generating data that can closely resemble human decisions made in the context of an interactive human-AI system like a computer game. We propose a novel algorithm that can generate synthetic, human-like, decision making data while starting from a very small set of decision making data collected from humans. Our proposed algorithm integrates the concept of reward shaping with an imitation learning algorithm to generate the synthetic data. We have validated our synthetic data generation technique by using the synthetically generated data as a surrogate for human interaction data to solve three sequential decision making tasks of increasing complexity within a small computer game-like setup. Different empirical and statistical analyses of our results show that the synthetically generated data can substitute the human data and perform the game-playing tasks almost indistinguishably, with very low divergence, from a human performing the same tasks.

\section{Introduction}

AI and machine learning (ML)-based systems are becoming increasingly  pervasive in the military for aiding the warfighter in making complex decisions. Recent examples of AI-assisted technology include AI-assisted fighter jet maneuvering, AI-enabled warfighter training in table-top, war-gaming exercises and AI-aided search and reconnaissance of dangerous and unsafe regions. In each of these application domains, it is essential to train the AI with human interaction data so that it can operate seamlessly when deployed alongside human warfigthers. Acquiring sufficient quantities of data of acceptable quality for ML training is a very challenging problem as the data needs to be well-structured, annotated or labeled, complete, reliable, and consistent. Most application domains, especially those of relevance to the military, in contrast, are characterized by sparse, un-labeled and at times incomplete data. To address this deficit, researchers have proposed techniques like data augmentation~\cite{shorten2019survey, feng2021survey} for synthetically generating data that can be used for ML training and testing. However, much of the existing research on data augmentation relates mostly to image data and, to a lesser extent to other data modalities like text, audio and graphical data. There has been relatively less research on synthetically generating human interaction data between a human and a machine. Acquiring data related to human actions and behavior also presents some additional challenges as human participants are difficult to recruit, once recruited they usually require considerable training to develop skills for the problem to be able to generate reliable data, and, finally, it is difficult to assure the quality of human-generated data without another human expert inspecting and validating the data. To alleviate these issues, it makes sense to investigate techniques that could synthetically generate data that closely replicates human decision making behavior while starting with a small set of human-generated decision making data.

In this paper, we propose a novel technique to generate synthetic data using imitation learning within the context of human-AI interactions in a computer game-playing scenario. The game playing scenarios require humans to solve sequential decision making tasks and the human interaction data in these tasks are represented as trajectories of game-state and human action pairs. The core of our approach uses a technique called reward shaping~\cite{Ng99} from reinforcement learning literature to train an agent from a small set of human decisions collected in the form of game-play data. This base set is then refined using an imitation learning algorithm called {\tt DAgger}~\cite{dagger} to create the final set of synthetically generated trajectory data. The synthetic data is validated in two ways. First, we use a divergence metric for trajectory data called METEOR ~\cite{banerjee2005meteor, choi2021trajgail} and show that the divergence score between the synthetic and original human-demonstrated data is within acceptable limits. As a stronger test, we substitute the synthetic data in place of human game-play data and evaluate if the game can be completed successfully with performance that is comparable to the human plays. Our results show that with the synthetic data the tasks in game can be completed $100 \%$ of the time with the similarity to the human plays increasing when we extend the reward shaping with imitation learning on the human data. This provides strong validation that the synthetic data retains the features of the human decision making data during game-plays. To the best of our knowledge, our paper is one of the first attempts at synthetically generating human-like data for sequential decision making tasks.

\section{Related Work}
 
{\bf Data Augmentation and Synthetic Data.} Data augmentation techniques use generative adversarial networks (GANs) at their core. A widely used GAN framework for data augmentation is the conditional GAN (cGAN)~\cite{mirza2014conditional} where the generator can generate data with specific class labels. The cGAN concept has been extended with different techniques including multiple loss functions in the Auxiliary Conditional GAN (ACGAN)~\cite{odena2017conditional} and CycleGAN~\cite{zhu2017unpaired}, using an autoencoder as generator in Data Augmentation GAN (DAGAN)~\cite{antoniou2017data} and Balancing GAN (BAGAN)~\cite{mariani2018bagan}, multiple generators called Siamese networks (Siamese-cGAN)~\cite{mukhtar21sidechannel} to generate synthetic data. Most data augmentation research has been in the field of computer vision for generating synthetic image data using a base set sampled from public image datasets. Recently data augmentation techniques have been applied to other data modalities including medical and agricultural imagery, \cite{frid-adar2018gan_medical, motamed2021xray, lu2022ganagreview}, urban vehicle movement data~\cite{choi2021trajgail}, tabular relational data \cite{SDV, xu2019tabularcgan}, vibration data from a mechanical sensor~\cite{shao2019acgan} and cryptography attack data~\cite{mukhtar21sidechannel}. In contrast to the aforementioned data modalities, GAN-based data augmentation has been relatively less successful for data that has a temporal component such as natural language data~\cite{alvarez-melis2022ganNLP} and human expert-generated temporal data that is considered in this work. GAN-based data augmentation models also require non-trivial quantities of training data and are challenging to use in sparse data environments. 
 
Synthetic data generation is the process of replicating data with novel data whose statistical properties correspond to that of the original data and has applications in many fields. In the medical sector, synthetic data has been used to build models on patient health information when unavailable because of privacy protections~\cite{hernandez22syn_health, torfi22synmedical}, and for analyzing disease risk factors~\cite{chen22disease_risk}. In the industrial sciences, synthetic data has been used to generate test burst data to prevent corrosion in metal pipes~\cite{he22burst_pipe}, simulating electrical load data~\cite{yilmaz22elecload}, government sector traffic volume forecasting~\cite{zhu2022traffic}, and rainwater synthetics for flood prediction~\cite{welten2022rainfall}. Most of these models generate continuous values in contrast to our problem of synthetic generation of nominal time-series data.

{\bf Reward Shaping.} Reward shaping updates the reward at a state-action pair using a value that is a function of the current state-action and future state-action pairs~\cite{grzes17rewardshaping}. One of the earliest works on reward shaping~\cite{Ng99} proposed this function as the difference of a potential function between the next and current states, which was approximated using the negative Manhattan distance to the goal between the two states. Subsequent researchers have extended this idea by proposing variations to the potential function using future as well as past states~\cite{Wiewiora03}, using distance to nearest states given in expert demonstrations~\cite{Brys15}, and using a bi-level optimization problem to simultaneously determine optimal values for the potential function and policy parameter (e.g., weights of a policy network) of the learning problem~\cite{hu2020learning}. Our work is fundamentally different from these techniques as their objective is to modify rewards so that the policy converges faster while our objective is to train a policy that could then be used to generate trajectories that are similar to human-demonstrated trajectories.

\section{Mathematical Framework for Generating Synthetic Trajectories}
We use a reinforcement learning (RL)-based technique that is formalized as a Markov Decision Process (MDP) ${\cal M} = (S, A, T, R, \gamma)$ as the framework for our problem. $S$ is the set of states in the environment, $A$ is the set of actions that a human or agent could take, $T: S \times A \times S \rightarrow [0, 1]$ is a transition model that specifies the forward dynamics of the environment, $R: S \times A \rightarrow \Re$ is the reward function and $\gamma \in [0, 1]$ is a discount factor. A sequence of state and action pairings is called a trajectory. The $i$-th trajectory is denoted by $\tau_i = (s_{i,k}, a_{i,k})_{k=0}^{|\tau_i|}$. 

We use $\mathcal{H} = \{\tau^h\}$ to denote the set of human demonstrated trajectories, where each $\tau_j^h = \{(s_j^h, a_j^h)\}$. For notation convenience, we denote $(S^h, A^h):S^h \subseteq S, A^h \subseteq A$, as the set of human-demonstrated states and actions respectively.

We have divided our approach into two steps. In the first step, we train an agent using deep RL and reward shaping that can generate trajectories similar to the human demonstrated trajectories. In the second step, we use imitation learning to generate synthetic trajectories while considering the agent created in the first step as a surrogate of a human expert that is generating trajectories. We describe these two steps in more details below:

\subsection{Training Expert Agent from Human-Generated Trajectories via Reward Shaping}
Our objective in this step is to create an expert agent that can learn to generate trajectories that are similar to the human-generated trajectories. Mathematically, the policy learned by the expert agent should preserve the distribution of the state-action pairs in the human-generated trajectories. This is a non-trivial problem as there only very few human-generated trajectories that cover only a fraction of the states in the state space. To address this problem, we use the reward shaping~\cite{Ng99} technique where the rewards at state-action pairs in the trajectories generated by the expert agent are updated in proportion to their divergence from a state-action pair on a human-demonstrated trajectory using Equation~\ref{eqn:f_value}.

\begin{equation}
\label{eqn:f_value}
\begin{aligned}
F (s, a, s')= 
    \begin{cases}
        \delta(s), & \text{if } \exists (s^h, a^h) \in (S^h, A^h): (s, a) = (s^h, a^h)\\
        0, & \text{if } \exists (s^h, a^h) \in (S^h, A^h): s = s^h, a \neq a^h\\
        \gamma\phi(s')-\phi(s), & \text{otherwise }\\
    \end{cases}
    \end{aligned}
\end{equation} 
\noindent
where, $s^G$ is a target or goal state in the problem, $\delta(s) = 1 - \frac{D_{min}(s, s^{G})}{D_{max}(s, s^{G})}$, $\phi(s)=1-\frac{D_{min}(s, s^h)}{D_{max}(s,s^h)}, s^h \in S^h$, $D_{min}, D_{max}$ are the shortest and longest paths between two states, $\phi(s)$ is a potential function that returns the suitability of state $s$ towards reaching the goal, and, $\gamma$ is a discount factor.

The logic behind the reward shaping equation is the following: 
If a state-action pair $(s, a)$ reached by the agent matched a state-action pair $(s^h, a^h)$from the human demonstrated trajectory, then $F$ was calculated by normalizing the shortest path from the current state $s$ to the goal state $s^G$. The updated reward at state $s$ gives the agent a higher incentive to take the same action as was demonstrated by a human.
If only the agent's current state matched a state in the human-demonstrated trajectories but not the agent's action, then the $F$ value was $0$, that is, the reward at state $s$ is left unchanged. This allows the agent to continue exploring actions at $s$ while using the initial reward distribution. Finally, if the agent's state $s$ was not in the set of human demonstrated trajectories $\mathcal{H}$, the $F$ value is given by a potential $\gamma\phi(s')-\phi(s)$ similar to~\cite{Ng99} where the potential function $\phi(s)$, is a normalized distance to the shortest path length from the agent's current state $s$ to the nearest state $s^h$ in the set of human demonstrated trajectories and $s'=\arg \max_{\hat{s}} T(s, a, \hat{s})$ is the next state returned by the transition function by taking action $a$ at state $s$. The updated reward at state $s$ then incentivizes the agent towards the goal and keeps the agent on track to reach the goal.

Algorithm~\ref{algo_train_expert} gives the algorithm for training the expert agent using reward shaping from human trajectories. The agent repeatedly generates trajectories using its current policy, applies the reward shaping function in Equation~\ref{eqn:f_value} to the states in the generated trajectories and update the policy using the shaped rewards. During training iteration $k$, the agent's nominal reward $R(s, a)$ is augmented as $R(s, a) \leftarrow R(s, a) + F(s, s')$ where $s'=\arg \max_{\hat{s}} T(s, a, \hat{s})$. The agent is considered as an expert when it can consistently generate trajectories with an average length above $N_{thresh}$ over a window size $W$.

\begin{algorithm}[htb!]
\SetAlgoLined
\SetKwInOut{Input}{input}\SetKwInOut{Output}{output}
\SetKwProg{myproc}{Procedure}{}{}
\Input{$S, A, T, R, \gamma$: MDP underlying environment\\
$N_{thresh}$: Mean episode length threshold\\
$T_{demo}$: Set of human expert trajectories\\}
\Output{$\pi_{RL}$: expert agent policy}
\BlankLine
\SetKwFunction{proc}{Train-Expert-Agent}{}
\myproc{\proc{$S, A, T, R, \gamma$, $N_{thresh}$, $RL_{algo}$, $T_{demo}$}}{
$k \leftarrow 0$\\
\While{($\frac{\sum^k_{k-W} len(\tau_k)}{W} < N_{thresh}$)} {
    $\tau_k \leftarrow$ Trajectory generated using current policy $\pi^{RL}_k$\\ 
    \For{each $(s, a) \in \tau$}{
        $R(s, a) \leftarrow R(s, a) + F(s, a, s')$, where $F(s, a, s')$ is given by Equation~\ref{eqn:f_value}\\
    }
    $\pi^{RL}_{k+1} \leftarrow$ update policy $\pi^{RL}_k$ with shaped rewards\\ 
    $k \leftarrow k+1$\\
    }
return $\pi_{RL}$ 
}
\caption{Training expert agent from human-generated trajectories.}
\label{algo_train_expert}
\end{algorithm}

\subsection{Generating Synthetic Trajectories via Imitation\\
Learning}

\begin{algorithm}[htb!]
\SetAlgoLined
\SetKwInOut{Input}{input}\SetKwInOut{Output}{output}
\SetKwProg{myproc}{Procedure}{}{}
\Input{$S, A, T, R, \gamma$: MDP underlying environment\\
$N_{train}$: {\tt DAgger} training length\\
$\pi^{RL}$: Expert agent policy\\}
\Output{$T_{syn}$: set of synthetically generated trajectories}
\BlankLine
\SetKwFunction{proc}{Generate-Synthetic-Traj}{}
\myproc{\proc{$S, A, T, R, \gamma$, $N_{train}$, $\pi{RL}$}}{
$\pi^{RL} \leftarrow $ {\tt generate-expert-agent()} \\
$\mathcal{D} \leftarrow T_{demo}$ \\
$\hat{\pi}_1 \leftarrow \pi^{RL}$ \\
\For{$i=1$ \bf{to} $N_{train}$}{
$\pi_i \leftarrow \beta_i\pi_{RL} + (1-\beta_i)\hat{\pi}_i$ \\
//Sample a trajectory using $\pi_i$\\
$\tau^{\pi_i} \leftarrow (s_0,\pi_i(s_0), ..., s_{G-1}, \pi_i(s_{G=1}), s_G)$\\
//Get actions from expert policy for states visited in $\tau^{\pi_i}$ \\
${\mathcal D_i} \leftarrow \emptyset$\\
\For{every $s_j \in \tau^{\pi_i}$}{
    ${\mathcal D_i} \leftarrow {\mathcal D_i} \cup (s_j, \pi^{RL}(s_j))$
    }
$\mathcal{D} \leftarrow \mathcal{D} \cup \mathcal{D}_i$ \\
Train classifier $\hat{\pi}_{i+1}$ on $\mathcal{D}$\\}
${\pi^*} \leftarrow$ Policy giving highest return in validation\\
$T^{syn} \leftarrow (s_1, \pi^*(s_1), ..., s_H, \pi^*(s_H))$ \\
\textbf{return} $T^{syn}$ \\
}
\caption{Synthetic trajectory creation from human expert-generated trajectories}
\label{algo_syn_traj}
\end{algorithm}

After creating the expert agent, we leverage it to generate synthetic trajectories via imitation learning. Our imitation learning technique is based on the Dataset Aggregation ({\tt DAgger}) algorithm~\cite{dagger}, as given in Algorithm~\ref{algo_syn_traj}.   The main idea of the {\tt DAgger} algorithm~\cite{dagger} is to generalize the expert policy $\pi^{RL}$ to states that might not have been visited by it before. To do this, the algorithm mixes the expert policy with a policy constructed by training a classifier on states visited by the expert policy, using mixing parameter $\beta_i$ (line $6)$. Trajectories generated by this mixed policy are more likely than the expert policy to visit states that were previously not visited by the expert policy (line $8$). To reduce large variances from the expert policy's actions, the actions at the states visited by the mixed policy trajectory are circumscribed to follow the expert policy (lines $9-13$). The resulting trajectories are used to train the classifier for mixing at the next iteration (lines $14-15$). The best policy from the classifier is then assigned to $\pi^*$, and is sampled until the goal or the time horizon is reached, and assigned to $T^{syn}$. This trajectory is then returned as a synthetically generated trajectory. This process is repeated to generate a desired number of synthetic trajectories.


\section{Methods and Experimental Settings}
We have validated our proposed synthetic trajectory generation technique to replicate human decisions that are taken successively while playing simple maze navigation-type computer games. Through our experiments, we plan to answer the following three research questions:
\begin{enumerate}
    \item Can DQN agents be trained using reward shaping from a sparse set of human-generated data to complete simple decision-making tasks at different levels of difficulty?
    \item Can synthetic trajectories with low divergence from human-generated trajectories be generated from human generated trajectories using the imitation learning technique in Algorithm~\ref{algo_syn_traj}?
    \item What is the effect of the imitation learning in Algorithm~\ref{algo_syn_traj}, used either without and with human data, on the quality of generated trajectories?
\end{enumerate}

\begin{figure}
\begin{tabular}{ccc}
    MAZE & CTF & CTFE \\
    \includegraphics[width=1.45in]{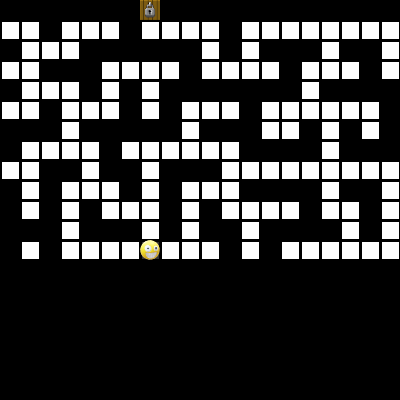} &  
     \includegraphics[width=1.45in]{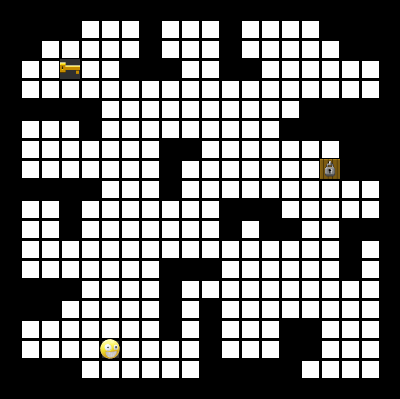} & 
     \includegraphics[width=1.45in]{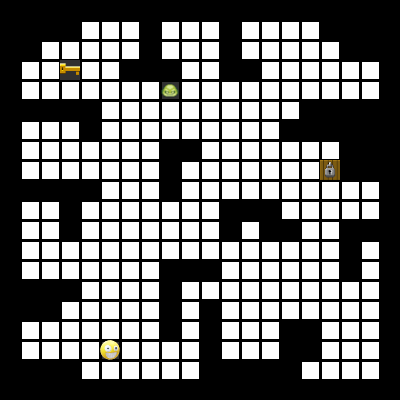}\\
     {\small (a)} & {\small (b)} & {\small (c)}
\end{tabular}
\caption{Maps of the different games used for experiments}
\label{fig:maps}
\end{figure}

\subsection{Game Environments}
We have used three maze navigation games of increasing complexities as scenarios for eliciting human decisions in our experiments. The games are used only as a means to elicit human decision made while moving a playing character in the game. We do not impose any criteria like rewarding players with higher scores if they choose better navigation routes while playing the game. Similarly, the agents used in our synthetic trajectory generation algorithm do not use any search or path planning algorithm to optimize their navigation path length while playing the game. The three games are described below:
\begin{itemize}
    \item \textbf{Maze Navigation Game.}  Figure~\ref{fig:maps}(a) illustrates the environment in a maze game (MAZE). The map consists of a $20 \times 13$ cell grid. Dark squares correspond to obstacles, light squares correspond to empty, navigable cells. The playing character (yellow face) is located in one of the empty cells towards the bottom of the map. The objective of the player is to navigate the playing character so that it reaches the goal location (brown cell with a lock icon). From its current cell, the playing character can be moved in one of the four cardinal directions: up, right, down, or left. If there is an empty cell in the direction of the movement, the playing character navigates into that cell. Otherwise, if it hits an obstacle in its direction of movement, it remains in its current cell. The obstacles are placed in a way so that there is only one route that leads from the start to the goal state; all other routes lead to dead ends. All routes are one cell wide. On reaching the goal, the player receives a score of $+1000$, while all intermediate moves had a score of $0$. The learning agent rewards were set to the same value as the human scores at the goal state. For non-goal states the agent's rewards are defined as a value that is proportional to the suitability of the state towards reaching the key or goal, as given by the equation below:
        \begin{equation}
    R =
    \begin{cases}
        100, & \text{if agent reaches goal}\\
        1 - \frac{D_{min}(s, s^{goal})}{D_{max}(s, s^{goal})}, & \text{otherwise} \nonumber \\
    \end{cases}    
    \label{eqn:maze_rewards}
    \end{equation}  
    \item \textbf{Capture-The-Flag (CTF) Game.} Figure~\ref{fig:maps}(b) shows the map used in our CTF game. The game is played in a $20 \times 20$ grid with obstacles to cause interference in human and agent movements, but not as constrained as the maze map.  Paths could be wider than $1$ cell width allowing areas that were more open for maneuvering around the map. The playing character is initially at the bottom left corner of the map, and the player's first task is to navigate the playing character through a maze to a key that is located near the top left corner of the map and collect it. After collecting the key, the player has to navigate to the goal (brown cell with a lock icon) near the middle of the map to finish the game. The player's score for collecting the key is $+100$, and for reaching the goal with the key is $+1000$.  Similar to the MAZE game, all intermediate moves have a score of $0$. For the learning agent, we set reward values based off the agent rewards structure in the MAZE game, but making the reward for non-goal states proportional to the agent's distance from its current objective (key or goal), as given by the equation below: 
    \begin{equation}
    R =
    \begin{cases}
        100, & \text{if agent collects key}\\
        1000, & \text{if agent reaches goal with key}\\
         1 - \frac{D_{min}(s, s^{key})}{D_{max}(s, s^{key})}, & \text{if agent does not have key}\\
        1 - \frac{D_{min}(s, s^{goal})}{D_{max}(s, s^{goal})}, & \text{if agent has key} \nonumber \\
    \end{cases}    
    \label{eqn:ctf_rewards}
    \end{equation}  
    \item \textbf{Capture-The-Flag With Enemy (CTFE) Game.} Figure~\ref{fig:maps}(c) illustrates the CTFE game. This game is identical to the CTF game, but with the addition of a roaming enemy that repeatedly moved in a horizontal, back-and-forth manner just below the key. If the player comes within $1$ cell distance to the enemy, it is captured by the enemy and loses the game with score $0$. For all other cases, the same player scores and agent reward values as in the CTF game were utilized. The agent reward function for the CTFE game is given below:
        \begin{equation}
    R =
    \begin{cases}
        100, & \text{if agent collects key}\\
        1000, & \text{if agent reaches goal with key}\\
         1 - \frac{D_{min}(s, s^{key})}{D_{max}(s, s^{key})}, & \text{if agent does not have key}\\
        1 - \frac{D_{min}(s, s^{goal})}{D_{max}(s, s^{goal})}, & \text{if agent has key}\\
        0, & \text{if captured by enemy} \nonumber \\
    \end{cases}    
    \label{eqn:ctfe_rewards}
    \end{equation}  
\end{itemize}

\subsection{Evaluation Metrics}
To evaluate the efficacy of our proposed technique, we have used the METEOR score~\cite{meteor}, METEOR was originally used a similarity measure for machine-based language translation and subsequently adapted as a distance measure between vehicle trajectories in autonomous driving~\cite{choi2021trajgail}. We selected the METEOR score as our metric of choice as it accounts for the order of the state-action pairings as well as their frequency of occurrence, as opposed to a one-to-one comparison metric. The METEOR score of a sentence translated from a reference language is given by:
\begin{equation}
    \text{METEOR score} = \frac{\# \text{mapped words in translation}}{\# \text{words in translation}} \times \frac{\# \text{mapped words in ref.}}{\# \text{words in reference}}
\label{eqn:meteor}
\end{equation} 

For our problem, a trajectory, either human-demonstrated or synthetic, is represented as a string by converting each state-action pairing into words of the form "{\tt observation-id, action-id}". A human demonstrated trajectory-string is considered as the reference, while the synthetic trajectory-string is considered as the translation. The METEOR score of a synthetic trajectory w.r.t a human-demonstrated trajectory is calculated using Equation~\ref{eqn:meteor}. METEOR scores lie in the interval $[0,1]$ with $0$ signifying no match and $1$ a perfect match. 

\subsection{Software and Hardware}
\noindent {\bf Software libraries.} We implemented the game environments using Open AI Gym~\cite{gym}. Stable Baselines 3~\cite{stable-baselines3} was used for implementing the DQN algorithm and Imitation library~\cite{imitation} for implementing the {\tt DAgger} algorithm. For calculating path lengths, $D_{min}$ and $D_{max}$ in Equations~\ref{eqn:f_value} and~\ref{eqn:ctf_rewards}, we used the all pair shortest paths with Dijkstra's algorithm on a graph representation of the environment, implemented via the NetworkX~\cite{networkx} library. Finally, we used the  NLTK Tooklkit~\cite{nltk} for implementing the METEOR score. 

\noindent \textbf{Hardware.} DQN agents were trained on a Google Collab server with $26$GB of RAM, four $2.30$ GHZ dual-core Intel Xeon CPUs, each with $46$MB cache, one NVIDIA Tesla T4 GPU with $16$GB of RAM, with Chromium $13$ as the operating system.  {\tt DAgger} algorithms were trained on a laptop with $8$GB RAM, four $2.1$ GHz dual-core AMD Ryzen with $5$ CPUs, each with $6$ MB cache and integrated AMD Radeon Vega-$8$ GPU running Microsoft Windows $10$ operating system. 

\omitit{
\textbf{Algorithms} \textit{DQN} \cite{dqn} A Deep Q Network (DQN) is a reinforcement learning algorithm based upon Q-learning.  It consists of a deep neural network that takes as input the current state and outputs an estimated expected cumulative reward that an agent can receive for each action that can be undertaken at the given state.  The network is trained by drawing random batches from an experience replay buffer, where state-action-state pairings are stored, and is updated by calculating the difference between the predicted Q values and the target Q values. 

\textit{{\tt DAgger}} \cite{{\tt DAgger}} Dataset Aggregation ({\tt DAgger}) is a reinforcement learning algorithm that is used to increase the performance of an existing policy.  It works by first gathering a dataset of trajectories by querying the expert policy.  Using this dataset it trains a new policy.  This policy is then sampled to generate a new trajectory.  Actions from the expert policy are then paired to each state visited in the trajectory, and the trajectory is aggregated into the dataset.  Then a supervised learning models is trained so that it will predict the actions the expert would most likely take given a specific state.  The policy learned from the supervised learning is then used in the next iteration, and the process repeats until an iteration count is met.  Upon completion the algorithm returns the best policy with the smallest amount of loss. 
}

\section{Experimental Results}
This section describes the results of the experiment across the three games, MAZE, CTF, CTFE, that were performed to answer the research questions outlined previously. We collected $5$ human trajectories for each game. The average length of the human trajectory for the different games is given in Table~\ref{table:parameters}. The different hyper-parameters for training the DQN agent in Algorithm~\ref{algo_train_expert} are given in Table~\ref{table:parameters}. Some of the hyper-parameter settings were the default values used by Stable-Baselines 3, while others were tuned to different settings to improve the performance of the DQN agents.  

\subsection{Can DQN Agents Be Trained Using Reward Shaping from a sparse set of human-generated data?}

\begin{table}[htb!]
    \centering
    \renewcommand{\arraystretch}{1.2}
    \begin{tabular}{|l|  m{1.75cm} |m{1.75cm} | m{1.75cm} |}
        \hline
         DQN hyper-parameter & MAZE & CTF & CTFE  \\
         \hline
         \hline
         Exploration Fraction & $0.8$ & $0.8$ & $0.99$ \\
         Exploration Initial EPS & $0.9$ & $0.9$ & $0.9$ \\
         Exploration Final EPS & $0.1$ & $0.1$ & $0.001$ \\
         $\gamma$ (discount factor) & $0.999$ & $0.999$ & $0.999$ \\
         Learning Starts & $1\times 10^5$ & $3\times 10^5$ & $3\times 10^5$ \\
         Learning Rate & $1 \times 10^{-4}$ & $1 \times 10^{-4}$ & $1 \times 10^{-4}$ \\
         Training Time-steps & $2 \times 10^6$ & $2 \times 10^6$ & $1.8 \times 10^6$ \\
         \hline
         Algo. input or parameter & \multicolumn{3}{|c|}{} \\
         \hline
         \hline
         No. human demo trajectories & $5$ & $5$ & $5$ \\
         Avg. human traj. length & $35$ & $35$ & $34$ \\
         $N_{thresh}$ (Algo. ~\ref{algo_train_expert}) & $55$ & $40$ & $42$ \\
         $W$ (Algo. ~\ref{algo_train_expert}) & $10$ & $10$ & $10$\\
         \hline
    \end{tabular}
    \caption{DQN algorithm hyper-parameters and Algorithms~\ref{algo_train_expert} and ~\ref{algo_syn_traj} parameters in the MAZE, CTF, and CTFE games.}
    \label{table:parameters}
\end{table}

To answer the first research question, three DQN agents were trained on three different maps: Maze, CTF, and CTFE. Reward shaping values derived from human demonstrator trajectories were used during training. The $N_{thresh}$ parameter value in Algorithm~\ref{algo_train_expert} was set to $55$ for the maze map, which is significantly higher than the $35$ steps it took the human to complete the map on average. Longer-than-average trajectory lengths were mostly due to oscillations of actions causing the agent to move back and forth to neighboring states or attempting repetitive sequences of actions within the same state.  However, this behavior was not demonstrated when the trained model was utilized to navigate the maze map during similarity evaluations.

\begin{figure}[htb!]
\begin{tabular}{cc}
\hspace*{-0.2in} \includegraphics[width=2.6in]{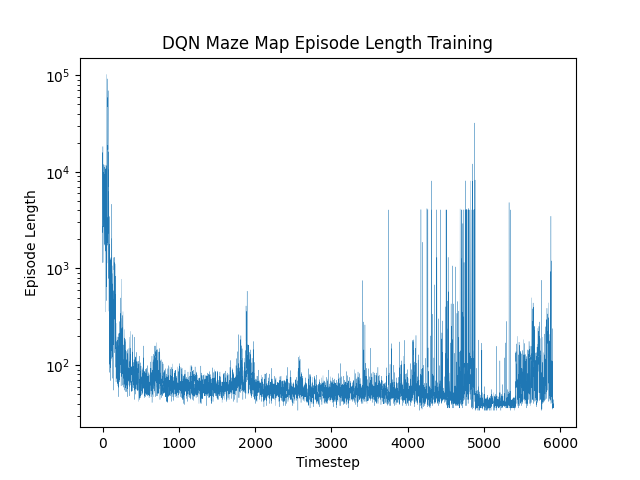}&
\hspace*{-0.4in} \includegraphics[width=2.6in]{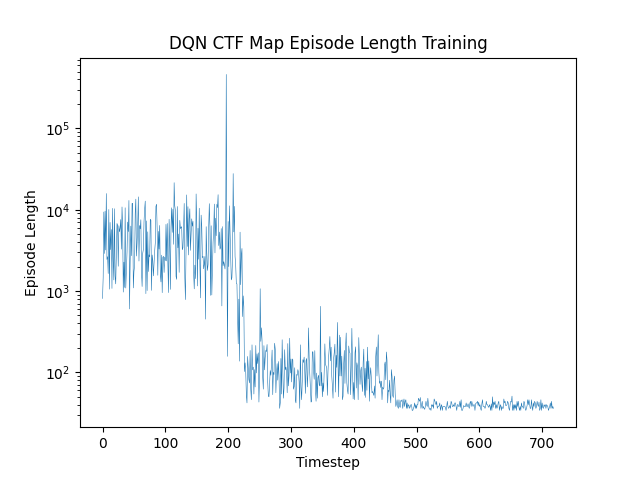}\\
{\small (a)} & {\small (b)}\\
\multicolumn{2}{c}{\includegraphics[width=2.6in]{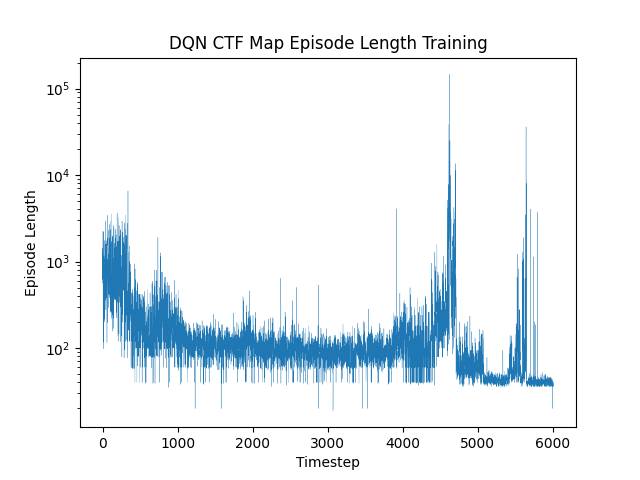}} \\
\multicolumn{2}{c}{\small (c)}
\end{tabular}
\caption{DQN episode length during training for different games.}
\label{fig:dqn_training}
\end{figure}

The results of episode length during training are visualized in Figure~\ref{fig:dqn_training}. All trained models were able to complete the game, with Maze DQN and CTFE DQN taking $6000$ time-steps to converge, while the CTF DQN took $720$ time-steps to converge.  This discrepancy could be explained by the differences in the map structure and task.  The maze map has very restricted movement with a greater chance of getting "stuck" in a corner, making it difficult to navigate away from.  Similarly, the CTFE map demonstrates an increase in difficulty with having a multi-point navigation task and the possibility of being destroyed by an enemy.  An agent could successfully navigate to the key and retrieve it but get destroyed by the enemy, effectively getting trapped into a limited state-action sequence similar to the maze map. Overall, the results indicate that DQN agents can be trained with reward shaping from a sparse set of human demonstrations for decisions taken in each of the three games.

\subsection{Can Low-Divergence Synthetic Trajectories be Generated from Human Demonstrated Trajectories?}
\begin{figure}[h!]
\begin{tabular}{cccc}
    &   DQN & {\tt DAgger}-E & {\tt DAgger}+E \\
    \vspace{-0.7in}
    {\rotatebox[origin=c]{90}{{\hspace*{1.0in} MAZE}}} & 
     \hspace{-0.2in} \includegraphics[width=1.7in]{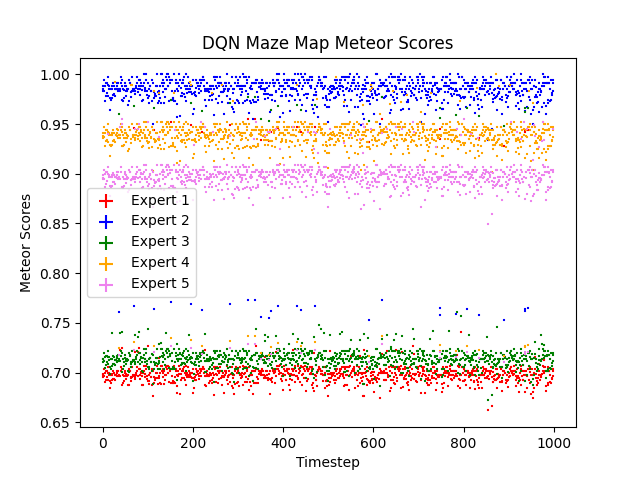} & 
     \hspace{-0.3in} \includegraphics[width=1.7in]{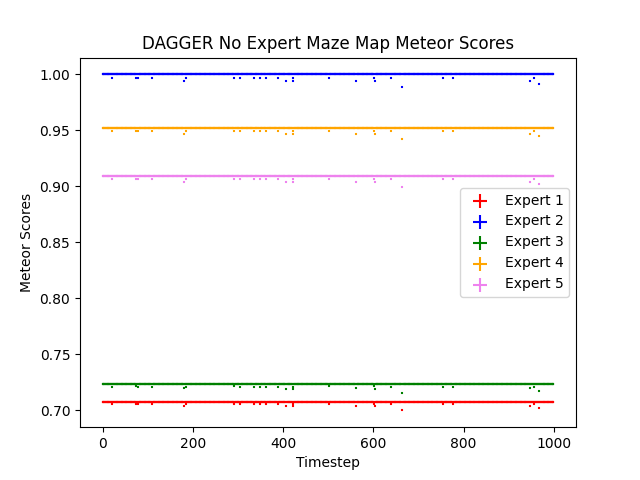} &
     \hspace{-0.4in} \includegraphics[width=1.7in]{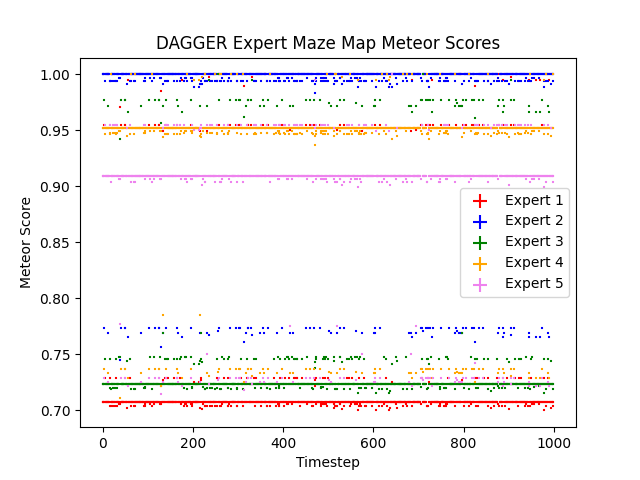}\\
     \vspace{-0.65in}
    \rotatebox[origin=c]{90}{\hspace*{1.0in} CTF} &
     \hspace{-0.2in} \includegraphics[width=1.7in]{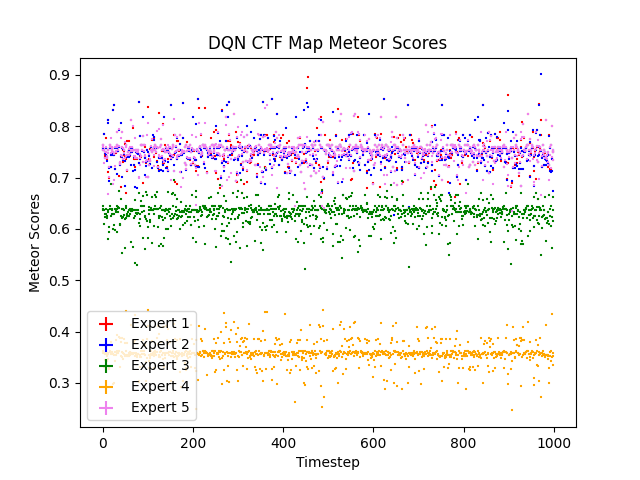}&
     \hspace{-0.3in} \includegraphics[width=1.7in]{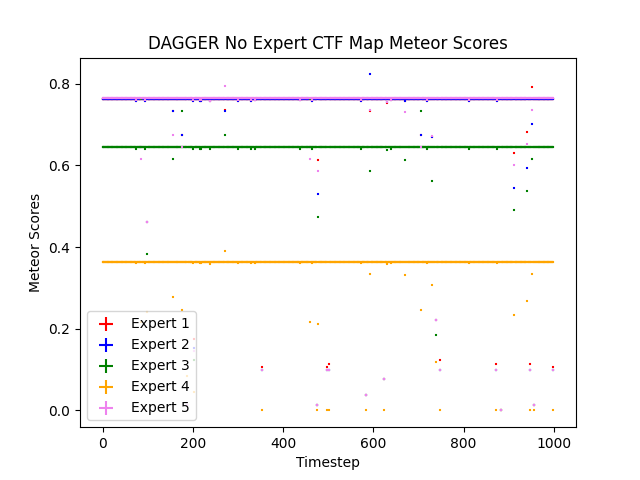}& 
     \hspace{-0.25in} \includegraphics[width=1.6in]{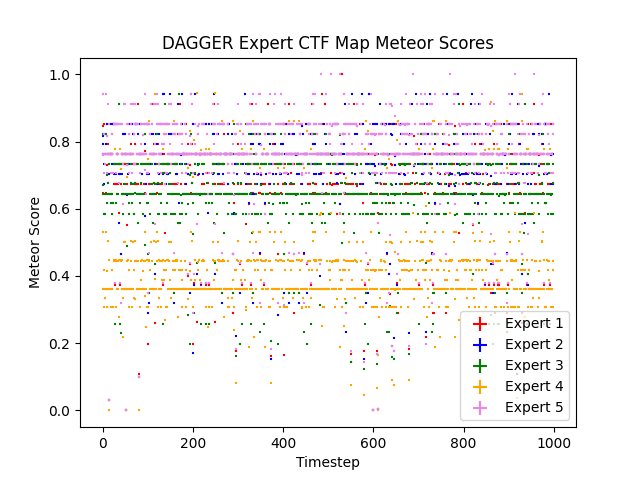}      \\
    \vspace{-0.7in}
    \rotatebox[origin=c]{90}{\hspace*{1.0in} CTFE} &
     \hspace{-0.2in} \includegraphics[width=1.7in]{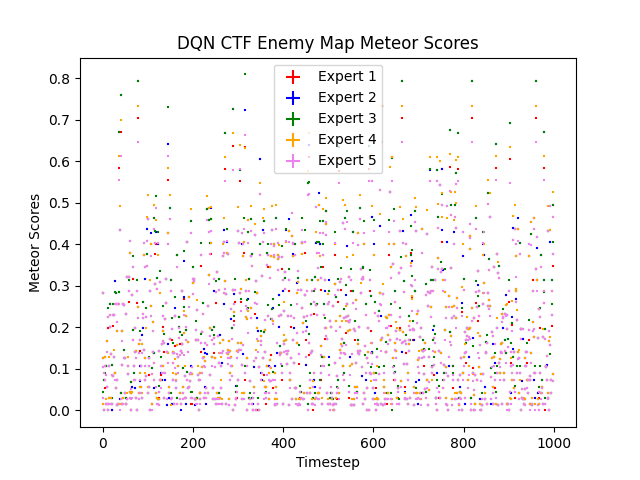}&
     \hspace{-0.4in} \includegraphics[width=1.7in]{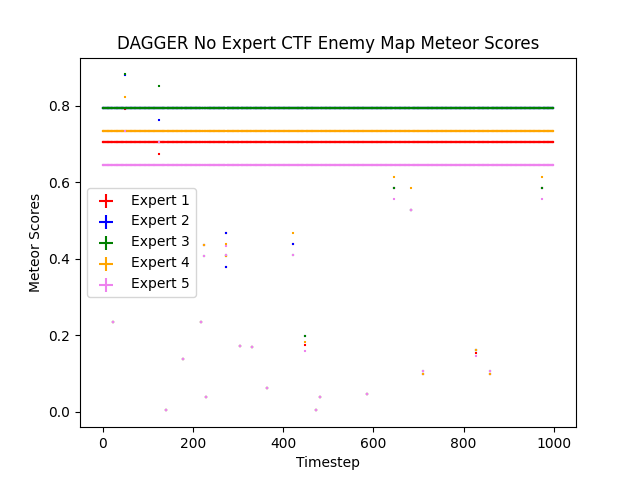} &
     \hspace{-0.4in} \includegraphics[width=1.7in]{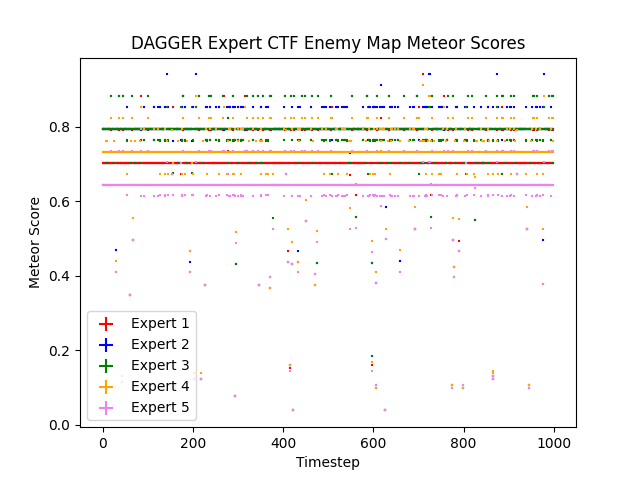}     \\
\end{tabular}
\caption{METEOR scores for all three games in our experiments.}
\label{fig:meteor_scores_games}
\end{figure}

To evaluate whether synthetic trajectories can be generated using only human demonstration trajectories, two different types of {\tt DAgger} algorithms were trained for each game. The first algorithm did not initially add human demonstrated trajectories to the dataset $\mathcal{D}$ as outlined on line $3$ of Algorithm~\ref{algo_syn_traj}. Instead, a trajectory set generated from the expert policy was used $\tau^{RL} \leftarrow \pi^{RL}$, $\mathcal{D}\leftarrow \tau^{RL}$.  This algorithm was referred to as {\tt DAgger}-E, or {\tt DAgger} with no expert trajectories. The second algorithm used the $5$ human demonstrator trajectories from reward shaping to train the DQN agents by adding them to the dataset $\mathcal{D}$ as outlined on line 3 of Algorithm ~\ref{algo_syn_traj}, and was refereed to as {\tt DAgger}+E, or, {\tt DAgger} with human demonstrator trajectories. After the completion of training, $1000$ trajectories were generated from each of the three games using DQN, {\tt DAgger}-E and {\tt DAgger}+E algorithms. These trajectories were used to calculate the METEOR scores against each of the $5$ human demontrated trajectories. Figure~\ref{fig:meteor_scores_games} shows the METEOR scores for each game for each algorithm and Table~\ref{table:meteor_scores} shows the average METEOR score over $1000$  generated trajectories.

\begin{table}[htb!]
\begin{center}
\renewcommand{\arraystretch}{1.2}
\begin{tabular}{| l | c c c c c |} 
  \hline
   Algorithm & Expert1 & Expert 2 & Expert 3 & Expert 4 & Expert 5 \\
  \hline
  \multicolumn{6}{|c|}{MAZE}\\
  \hline
  DQN & $0.70$ & $\mathbf{0.97}$ & $.72$ & $0.93$ & $0.89$\\
  {\tt DAgger}-E & $0.70$ & $\mathbf{0.99}$ & $0.72$ & $0.95$ & $0.90$\\
  {\tt DAgger}+E & $0.73$ & $\mathbf{0.97}$ & $0.75$ & $0.93$ & $0.89$\\
  \hline
  \multicolumn{6}{|c|}{CTF}\\
  \hline
  DQN & \textbf{0.75} & $0.74$ & $0.62$ & $0.35$ & $0.74$ \\
  {\tt DAgger}-E & $\textbf{0.75}$ & $\textbf{0.75}$ & $0.63$ & $0.35$ & $\textbf{0.75}$ \\
  {\tt DAgger}+E & $0.71$ & $0.73$ & $0.63$ & $0.41$ & $\textbf{0.75}$ \\
  \hline
  \multicolumn{6}{|c|}{CTFE}\\
  \hline
  DQN & $0.17$ & $0.18$ & $0.18$ & $0.18$ & $0.17$ \\
  {\tt DAgger}-E & $0.69$ & $\textbf{0.78}$ & $\textbf{0.78}$ & $0.72$ & $0.63$ \\
  {\tt DAgger}+E & $0.70$ & $\textbf{0.78}$ & $0.76$ & $0.72$ & $0.63$ \\
  \hline
\end{tabular}
\caption{Average METEOR Scores for each game and agent compared to human experts demonstrators 1-5.}
\label{table:meteor_scores}
\end{center}
\end{table}

The three agents on the MAZE game were most similar to human demonstrator trajectory $2$.  {\tt DAgger}-E had the highest average METEOR score of $0.99$, while DQN and {\tt DAgger}+E had an average score of $0.97$. For the CTF game, the results varied more with each algorithm replicating a different human trajectory.  DQN was most similar to human demonstrator $1$ with a score of $0.75$, while {\tt DAgger}+E was most similar to human demonstrator $5$, also with a score of $0.75$.  {\tt DAgger}-E was an exception, as it was most similar to human demonstrators $1$, $2$, and $5$ with a score of $0.75$.  The similarity scores were lower when compared to the scores from the MAZE game.  This is likely due to the less constrained layout of the map, which allows for more movement as opposed to the MAZE game that has limited maneuverability. For the CTFE game, the DQN was not able to replicate any of the human trajectories, as the highest METEOR score was $0.19$.  {\tt DAgger}-E was most similar to human demonstrator trajectories $2$ and $3$ while {\tt DAgger}+E was most similar to human demonstrator $2$, with both having scores of $0.78$. Overall, these findings support that synthetic trajectories with very low divergence  from human trajectories can be generated using the technique proposed in this paper. 

\subsection{What is the effect of the imitation learning, used either without and with human data, on the quality of generated trajectories?}
To evaluate the performance differences between DQN and {\tt DAgger} and to determine the significance of the inclusion of human trajectory data sets during synthetic trajectory generation, one-way ANOVA were run between the METEOR scores for the DQN, {\tt DAgger}-E, and {\tt DAgger}+E for each of the three maps.  The results of these tests are outlined in Table~\ref{table:meteor_pvalues_anova}. All the results indicate significant differences for each meteor score between DQN, {\tt DAgger}-E, an {\tt DAgger}+E on all three maps, except for the 3rd METEOR score on the CTF MAP.  
  
\begin{table}[htb!]
\begin{center}
\renewcommand{\arraystretch}{1.2}
\begin{tabular}{|l| c c c c c|} 
   \hline
   Game  & {\small METEOR 1} & {\small METEOR 2} & {\small METEOR 3} & {\small METEOR 4} & {\small METEOR 5} \\
  \hline
  MAZE  & $102.64$ & $85.18$ & $111.12$ & $66.90$ & $46.30$ \\
        & $(<.001)$ & $(<.001)$ & $(<.001)$ & $(<.001)$ & $(<.001)$ \\
\hline
  CTF & $40.12$ & $5.48$ & $1.76$ & $175.36$ & $2.93$\\
       & $(<.001)$ & $(<.01)$ & $(0.17)$ & $(<.001)$ & $(<.05)$\\
    \hline
  CTFE & $7505$ & $8137$ & $7852$ & $7504$ & $6474$\\
       & $(<.001)$ & $(<.001)$ & $(<.001)$ & $(<.001)$ & $(<.001)$\\
  \hline
\end{tabular}
\caption{P Values for One-Way T-Tests and ANOVA}
\label{table:meteor_pvalues_anova}
\end{center}
\end{table}

To further understand the significance of the variances between and within all groups, we counted the frequencies of {\tt DAgger}-E METEOR scores that were above the average DQN METEOR score across all human demonstrator trajectories.  Additionally, we counted the frequencies of {\tt DAgger}+E scores that were above the average METEOR scores of both DQN and {\tt DAgger}-E.  These results are illustrated in Table~\ref{table:meteor_scores_compare}. Out of a total of $1000$ trajectories, {\tt DAgger}-E and {\tt DAgger}+E had a significant number of METEOR scores for each human demonstrator above the average DQN score, suggesting that while DQN could synthetically generate trajectories that were similar to human trajectories, {\tt DAgger} algorithms could synthetically generate more similar trajectories than DQN could.  Conversely, when comparing {\tt DAgger}+E and {\tt DAgger}-E frequency counts, {\tt DAgger}+E had a large number of scores that were greater than the average score of {\tt DAgger}-E. This suggests that incorporating human expert trajectories within the {\tt DAgger} algorithm increased the similarity to human demonstrator trajectories further and because these frequency counts were observed across all human demonstrator trajectories, meaning synthetic trajectories were more diversified in which human trajectory the algorithm replicated.  

\begin{table}[htb!]
\begin{center}
\renewcommand{\arraystretch}{1.2}
\begin{tabular}{| l  c c c |} 
  \hline
   & {\tt DAgger}-E & {\tt DAgger}+E & {\tt DAgger}+E   \\ 
   & above & above & above \\
   & DQN Avg. & DQN Avg. & {\tt DAgger}-E \\
  \hline
  \multicolumn{4}{|c|}{MAZE}\\
  \hline
  \hspace{1em}Expert 1 & 992 & 903 & 856 \\
  \hspace{1em}Expert 2 & 1000 & 901 & 696 \\
  \hspace{1em}Expert 3 & 992 & 913 & 901 \\
  \hspace{1em}Expert 4 & 1000 & 901 & 736 \\
  \hspace{1em}Expert 5 & 1000 & 901 & 809 \\
  \hline
  \multicolumn{4}{|c|}{CTF}\\
  \hline
  \hspace{1em}Expert 1 & 973 & 592 & 592 \\
  \hspace{1em}Expert 2 & 973 & 684 & 684 \\
  \hspace{1em}Expert 3 & 976 & 719 & 719 \\
  \hspace{1em}Expert 4 & 972 & 848 & 848 \\
  \hspace{1em}Expert 5 & 972 & 778 & 778 \\
  \hline
  \multicolumn{4}{|c|}{CTFE}\\
  \hline
  \hspace{1em}Expert 1 & 987 & 984 & 954 \\
  \hspace{1em}Expert 2 & 988 & 984 & 895 \\
  \hspace{1em}Expert 3 & 988 & 984 & 793 \\
  \hspace{1em}Expert 4 & 987 & 984 & 911 \\
  \hspace{1em}Expert 5 & 987 & 984 & 847 \\
  \hline
\end{tabular}
\caption{Frequency of METEOR Scores above METEOR Average}
\label{table:meteor_scores_compare}
\end{center}
\end{table}

\section{Conclusions and Future Work}
In this paper, we have demonstrated a novel technique for generating synthetic data for human decision making data while starting with a very small set of human-demonstrated data samples. To the best of our knowledge, this is one of the first attempts at generating synthetic versions of human decision making data. The close correspondence between the synthetic data and actual human-generated data for the different decision making tasks in our experiments supports our claim that it could be used as a replacement of actual human data for training machine learning models when it might be difficult to acquire sufficient quantities of high-quality human-generated data. While our claim has been validated for decision making in navigation-like tasks, it would be essential to verify if it could be generalized to more complex decisions, such as those encountered in war-gaming training exercises. Another aspect of our approach is that we first need to train an agent from the sparse human-generated data before using it, enhanced with imitation learning, to generate the final synthetic data. The synthetic data generation process could be made faster and more lightweight if we could reduce the requirement of training an agent using RL as it requires access to a decision process like the transition function of an MDP or the forward mechanics model of a game engine. Integrating our proposed approach with techniques for synthetically generating time-series data~\cite{bandara2021improving} could provide promising direction towards reducing the dependency. Our work in this paper is a first step towards synthetically creating larger quantities of human-decision making data that could be used for a variety of purposes including warfighter training, building human assistants for commanders and decision makers, and detecting patterns of human decisions for analysing decision traits. We envisage that this work will lay the foundations of future work towards a more thorough understanding of the problems in replicating human-decisions and aiding humans to make better decisions.

\section{Credit Authorship Contribution Statement}
Dasgupta was responsible for supervising the research, conceptualizing the initial research ideas and setting the general direction of the research. Brandt was responsible for identifying and finalizing the research methods, implementing all the software, setting up and running the experiments, and, collecting and presenting the research results. Sections $1-2$ and $6$ are written by Dasgupta, Sections $3-4$ are written jointly by Brandt and Dasgupta, Section $5$ is written by Brandt.

\section{Acknowledgements}
This research was done as part of the project {\em Playing Games to Overcome Cognitive Biases in Warfighter Decision Making} that is supported by a NRL Base Funding 6.1 grant from the Office of Naval Research. Bryan Brandt worked on this project as part of his NREIP summer internship in 2022 and subsequently as a student volunteer.

\bibliographystyle{plain}
\bibliography{synthetic_traj_reward_shaping}

\begin{thebibliography}{10}

\bibitem{alvarez-melis2022ganNLP}
David Alvarez-Melis, Vikas Garg, and Adam~Tauman Kalai.
\newblock Why gans are overkill for nlp, 2022.

\bibitem{antoniou2017data}
Antreas Antoniou, Amos Storkey, and Harrison Edwards.
\newblock Data augmentation generative adversarial networks.
\newblock {\em arXiv preprint arXiv:1711.04340}, 2017.

\bibitem{bandara2021improving}
Kasun Bandara, Hansika Hewamalage, Yuan-Hao Liu, Yanfei Kang, and Christoph
  Bergmeir.
\newblock Improving the accuracy of global forecasting models using time series
  data augmentation.
\newblock {\em Pattern Recognition}, 120:108148, 2021.

\bibitem{banerjee2005meteor}
Satanjeev Banerjee and Alon Lavie.
\newblock {METEOR}: An automatic metric for {MT} evaluation with improved
  correlation with human judgments.
\newblock In {\em Proceedings of the {ACL} Workshop on Intrinsic and Extrinsic
  Evaluation Measures for Machine Translation and/or Summarization}, pages
  65--72, Ann Arbor, Michigan, June 2005. Association for Computational
  Linguistics.

\bibitem{nltk}
Biard, Steven, Edward Loper, and Ewan Klein.
\newblock {\em Natural Language Processing with Python}.
\newblock O'Reilly Media Inc, 2009.

\bibitem{gym}
Greg Brockman, Vicki Cheung, Ludwig Pettersson, Jonas Schneider, John Schulman,
  Jie Tang, and Wojciech Zaremba.
\newblock Openai gym, 2016.

\bibitem{Brys15}
Tim Brys, Anna Harutyunyan, Halit~Bener Suay, Sonia Chernova, Matthew~E.
  Taylor, and Ann Now\'{e}.
\newblock Reinforcement learning from demonstration through shaping.
\newblock In {\em Proceedings of the 24th International Conference on
  Artificial Intelligence}, IJCAI'15, page 3352–3358. AAAI Press, 2015.

\bibitem{chen22disease_risk}
Anjun Chen.
\newblock A novel graph methodology for analyzing disease risk factor
  distribution using synthetic patient data.
\newblock {\em Healthcare Analytics}, 2:100084, 2022.

\bibitem{choi2021trajgail}
Seongjin Choi, Jiwon Kim, and Hwasoo Yeo.
\newblock Trajgail: Generating urban vehicle trajectories using generative
  adversarial imitation learning.
\newblock {\em Transportation Research Part C: Emerging Technologies},
  128:103091, 2021.

\bibitem{feng2021survey}
Steven~Y Feng, Varun Gangal, Jason Wei, Sarath Chandar, Soroush Vosoughi,
  Teruko Mitamura, and Eduard Hovy.
\newblock A survey of data augmentation approaches for nlp.
\newblock {\em arXiv preprint arXiv:2105.03075}, 2021.

\bibitem{frid-adar2018gan_medical}
Maayan Frid-Adar, Idit Diamant, Eyal Klang, Michal Amitai, Jacob Goldberger,
  and Hayit Greenspan.
\newblock Gan-based synthetic medical image augmentation for increased cnn
  performance in liver lesion classification.
\newblock {\em Neurocomputing}, 321:321--331, 2018.

\bibitem{imitation}
Adam Gleave, Mohammad Taufeeque, Juan Rocamonde, Erik Jenner, Steven~H. Wang,
  Sam Toyer, Maximilian Ernestus, Nora Belrose, Scott Emmons, and Stuart
  Russell.
\newblock imitation: Clean imitation learning implementations.
\newblock arXiv:2211.11972v1 [cs.LG], 2022.

\bibitem{grzes17rewardshaping}
Marek Grzes.
\newblock Reward shaping in episodic reinforcement learning.
\newblock In Kate Larson, Michael Winikoff, Sanmay Das, and Edmund~H. Durfee,
  editors, {\em Proceedings of the 16th Conference on Autonomous Agents and
  MultiAgent Systems, {AAMAS} 2017, S{\~{a}}o Paulo, Brazil, May 8-12, 2017},
  pages 565--573. {ACM}, 2017.

\bibitem{networkx}
Aric~A. Hagberg, Daniel~A. Schult, and Pieter~J. Swart.
\newblock Exploring network structure, dynamics, and function using networkx.
\newblock In Ga\"el Varoquaux, Travis Vaught, and Jarrod Millman, editors, {\em
  Proceedings of the 7th Python in Science Conference}, pages 11 -- 15,
  Pasadena, CA USA, 2008.

\bibitem{he22burst_pipe}
Z.~He and W.~Zhou.
\newblock Generation of synthetic full-scale burst test data for corroded
  pipelines using the tabular generative adversarial network.
\newblock {\em Engineering Applications of Artificial Intelligence},
  115:105308, 2022.

\bibitem{hernandez22syn_health}
Mikel Hernandez, Gorka Epelde, Ane Alberdi, Rodrigo Cilla, and Debbie Rankin.
\newblock Synthetic data generation for tabular health records: A systematic
  review.
\newblock {\em Neurocomputing}, 493:28–45, 2022.

\bibitem{hu2020learning}
Yujing Hu, Weixun Wang, Hangtian Jia, Yixiang Wang, Yingfeng Chen, Jianye Hao,
  Feng Wu, and Changjie Fan.
\newblock Learning to utilize shaping rewards: A new approach of reward
  shaping.
\newblock {\em Advances in Neural Information Processing Systems},
  33:15931--15941, 2020.

\bibitem{meteor}
Alon Lavie and Abhaya Agarwal.
\newblock Meteor: An automatic metric for mt evaluation with high levels of
  correlation with human judgments.
\newblock In {\em Proceedings of the Second Workshop on Statistical Machine
  Translation}, StatMT '07, page 228–231, USA, 2007. Association for
  Computational Linguistics.

\bibitem{lu2022ganagreview}
Yuzhen Lu, Dong Chen, Ebenezer Olaniyi, and Yanbo Huang.
\newblock Generative adversarial networks (gans) for image augmentation in
  agriculture: A systematic review.
\newblock {\em Computers and Electronics in Agriculture}, 200:107208, 2022.

\bibitem{mariani2018bagan}
Giovanni Mariani, Florian Scheidegger, Roxana Istrate, Costas Bekas, and
  Cristiano Malossi.
\newblock Bagan: Data augmentation with balancing gan.
\newblock {\em arXiv preprint arXiv:1803.09655}, 2018.

\bibitem{mirza2014conditional}
Mehdi Mirza and Simon Osindero.
\newblock Conditional generative adversarial nets.
\newblock {\em arXiv preprint arXiv:1411.1784}, 2014.

\bibitem{motamed2021xray}
Saman Motamed, Patrik Rogalla, and Farzad Khalvati.
\newblock Data augmentation using generative adversarial networks (gans) for
  gan-based detection of pneumonia and covid-19 in chest x-ray images.
\newblock {\em Informatics in Medicine Unlocked}, 27:100779, 2021.

\bibitem{mukhtar21sidechannel}
Naila Mukhtar, Lejla Batina, Stjepan Picek, and Yinan Kong.
\newblock Fake it till you make it: Data augmentation using generative
  adversarial networks for all the crypto you need on small devices.
\newblock Cryptology ePrint Archive, Paper 2021/991, 2021.
\newblock \url{https://eprint.iacr.org/2021/991}.

\bibitem{Ng99}
A.~Ng, Daishi Harada, and Stuart~J. Russell.
\newblock Policy invariance under reward transformations: Theory and
  application to reward shaping.
\newblock In {\em ICML}, 1999.

\bibitem{odena2017conditional}
Augustus Odena, Christopher Olah, and Jonathon Shlens.
\newblock Conditional image synthesis with auxiliary classifier gans.
\newblock In {\em International conference on machine learning}, pages
  2642--2651. PMLR, 2017.

\bibitem{SDV}
N.~{Patki}, R.~{Wedge}, and K.~{Veeramachaneni}.
\newblock The synthetic data vault.
\newblock In {\em 2016 IEEE International Conference on Data Science and
  Advanced Analytics (DSAA)}, pages 399--410, 10 2016.

\bibitem{stable-baselines3}
Antonin Raffin, Ashley Hill, Adam Gleave, Anssi Kanervisto, Maximilian
  Ernestus, and Noah Dormann.
\newblock Stable-baselines3: Reliable reinforcement learning implementations.
\newblock {\em Journal of Machine Learning Research}, 22(268):1--8, 2021.

\bibitem{dagger}
Stephane Ross, Geoffrey~J. Gordon, and J.~Andrew Bagnell.
\newblock A reduction of imitation learning and structured prediction to
  no-regret online learning, 2010.

\bibitem{shao2019acgan}
Siyu Shao, Pu~Wang, and Ruqiang Yan.
\newblock Generative adversarial networks for data augmentation in machine
  fault diagnosis.
\newblock {\em Computers in Industry}, 106:85--93, 2019.

\bibitem{shorten2019survey}
Connor Shorten and Taghi~M Khoshgoftaar.
\newblock A survey on image data augmentation for deep learning.
\newblock {\em Journal of big data}, 6(1):1--48, 2019.

\bibitem{torfi22synmedical}
Amirsina Torfi, Edward~A. Fox, and Chandan~K. Reddy.
\newblock Differentially private synthetic medical data generation using
  convolutional gans.
\newblock {\em Information Sciences}, 586:485–500, 2022.

\bibitem{welten2022rainfall}
Sascha Welten, Adrian Holt, Julian Hofmann, Lennart Schelter, Elena-Maria
  Klopries, Thomas Wintgens, and Stefan Decker.
\newblock Synthetic rainfall data generator development through decentralised
  model training.
\newblock {\em Journal of Hydrology}, 612:128210, 2022.

\bibitem{Wiewiora03}
Eric Wiewiora, Garrison Cottrell, and Charles Elkan.
\newblock Principled methods for advising reinforcement learning agents.
\newblock In {\em Proceedings of the Twentieth International Conference on
  International Conference on Machine Learning}, ICML'03, page 792–799. AAAI
  Press, 2003.

\bibitem{xu2019tabularcgan}
Lei Xu, Maria Skoularidou, Alfredo Cuesta-Infante, and Kalyan Veeramachaneni.
\newblock Modeling tabular data using conditional gan.
\newblock In {\em Advances in Neural Information Processing Systems}, 2019.

\bibitem{yilmaz22elecload}
Bilgi Yilmaz and Ralf Korn.
\newblock Synthetic demand data generation for individual electricity consumers
  : Generative adversarial networks (gans).
\newblock {\em Energy and AI}, 9:100161, 2022.

\bibitem{zhu2017unpaired}
Jun-Yan Zhu, Taesung Park, Phillip Isola, and Alexei~A Efros.
\newblock Unpaired image-to-image translation using cycle-consistent
  adversarial networks.
\newblock In {\em Proceedings of the IEEE international conference on computer
  vision}, pages 2223--2232, 2017.

\bibitem{zhu2022traffic}
Kun Zhu, Shuai Zhang, Jiusheng Li, Di~Zhou, Hua Dai, and Zeqian Hu.
\newblock Spatiotemporal multi-graph convolutional networks with synthetic data
  for traffic volume forecasting.
\newblock {\em Expert Systems with Applications}, 187:115992, 2022.

\end{thebibliography}

\end{document}